\documentclass[10pt,twocolumn,letterpaper]{article}

\usepackage[algorithms]{wacv}

\usepackage{times}
\usepackage{latexsym}

\definecolor{wacvblue}{rgb}{0.21,0.49,0.74}
\usepackage[pagebackref,breaklinks,colorlinks,allcolors=wacvblue]{hyperref}

\def\wacvPaperID{2515} %
\def\confName{WACV}
\def\confYear{2027}

\usepackage[T1]{fontenc}

\usepackage[utf8]{inputenc}

\usepackage{microtype}

\usepackage{inconsolata}

\usepackage{graphicx}
\usepackage{subcaption}
\usepackage{xspace}
\usepackage[table]{xcolor}
\usepackage{bm}
\usepackage{tcolorbox}
\usepackage{multicol}
\usepackage{enumitem}
\usepackage{multirow}
\usepackage{booktabs}
\usepackage[accsupp]{axessibility}
\usepackage{orcidlink}
\usepackage{wrapfig} 
\usepackage{amsmath}
\usepackage[accsupp]{axessibility}  %

\usepackage[capitalize]{cleveref}
\crefname{figure}{Fig.}{Figs.}
\Crefname{figure}{Fig.}{Figs.}

\crefname{section}{Sec.}{Secs.}
\Crefname{section}{Sec.}{Secs.}
\definecolor{cvprblue}{rgb}{0.21,0.49,0.74}
\definecolor{darkgreen}{rgb}{0.0, 0.5, 0.0}

\newcommand{\change}[1]{{\color{red}{#1}}}

\renewcommand{\change}[1]{#1}

\newcommand{\methodname}{SketchVerify\xspace}

\title{Verify Before You Generate: Sketch-Guided Efficient \\ Motion-Planning for Video Generation}

\author{
Yidong Huang$^{1}$ \quad
Zun Wang$^{1}$ \quad
Han Lin$^{1}$ \quad
Dong-Ki Kim$^{2}$ \quad
Shayegan Omidshafiei$^{2}$ \\
Jaehong Yoon$^{3}$ \quad
Yue Zhang$^{1}$ \quad
Mohit Bansal$^{1}$ \vspace{8pt} \\
$^{1}$UNC Chapel Hill \quad
$^{2}$ FieldAI \quad
$^{3}$ Nanyang Technological University \vspace{8pt} \\
{\tt \href{https://sketchverify.github.io/}{\textbf{https://sketchverify.github.io/}}}
}
\begin{document}
\maketitle
\begin{abstract}
Recent video generation approaches increasingly rely on planning intermediate control signals like object trajectories to improve temporal coherence and motion fidelity.
However, these methods mostly employ single-shot plans that are typically limited to simple motion instructions, leading to suboptimal motion quality,
or iterative refinement which requires multiple calls to the video generator and incurs high computational cost.
To address this trade-off, we propose \textbf{\methodname{}}, a training-free framework whose key insight is to verify motion plans using lightweight video sketches before invoking the expensive video generator.
This lightweight planning space enables efficient comparison of multiple candidate trajectories and bounded iterative refinement before a single final video-generation pass.
Given a prompt and a reference image, \methodname{} samples multiple candidate trajectories and renders each as a lightweight video sketch by compositing the moving objects over a static background. A vision-language verifier then ranks these sketches according to instruction alignment and visual motion consistency, allowing unsuitable trajectories to be rejected before full video synthesis. The motion plan
is iteratively refined until a satisfactory one is identified, then passed to a
trajectory-conditioned generator for final synthesis.  
Experiments on \textsc{WorldModelBench} and \textsc{PhyWorldBench} demonstrate consistent improvements in instruction following and temporally coherent, visually plausible motion.
Using the same ATI-14B generator, \methodname{} improves the aggregate WorldModelBench score from 7.71 to 8.71.
Compared with full-video iterative refinement, it achieves a 13.1$\times$ planning-stage speedup and a 3.6$\times$ end-to-end speedup.

\end{abstract}  
\section{Introduction}

Text-and-image-to-video (TI2V) generation has demonstrated strong potential across applications such as robotic manipulation, autonomous driving, and game content creation. While modern TI2V models~\cite{wan2025wan,singer2023make,chen2023videocrafter1,jang2025frame,ho2022imagen} achieve impressive visual fidelity and semantic alignment~\cite{bytedance_seed_2026}, generating precise and temporally coherent motion remains challenging. Models may animate the wrong object, follow an incorrect trajectory, miss a required interaction, or produce abrupt motion and implausible object interactions~\cite{gu2025phyworldbench,li2025worldmodelbench}. Such failures can persist even when individual frames appear realistic, making motion planning a central bottleneck for controllable video generation.

\begin{figure}[t]
  \centering
  \captionsetup[subfigure]{font=footnotesize,skip=3pt}
  \captionsetup{font=footnotesize,skip=3pt}

  \begin{subfigure}[t]{0.49\linewidth}
    \centering
    \includegraphics[width=\linewidth]{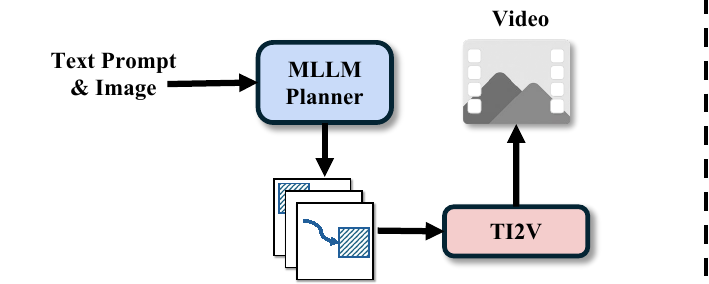}
    \caption{Single-shot planning.}
    \label{fig:teaser1}
  \end{subfigure}
  \hfill
  \begin{subfigure}[t]{0.49\linewidth}
    \centering
    \includegraphics[width=\linewidth]{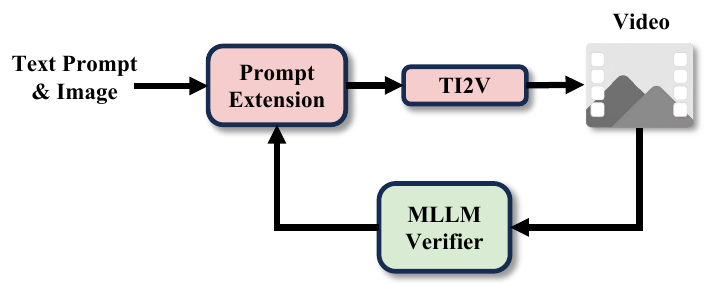}
    \caption{Iterative generation.}
    \label{fig:teaser2}
  \end{subfigure}

  \vspace{5pt}

  \begin{subfigure}[t]{0.98\linewidth}
    \centering
    \includegraphics[width=\linewidth]{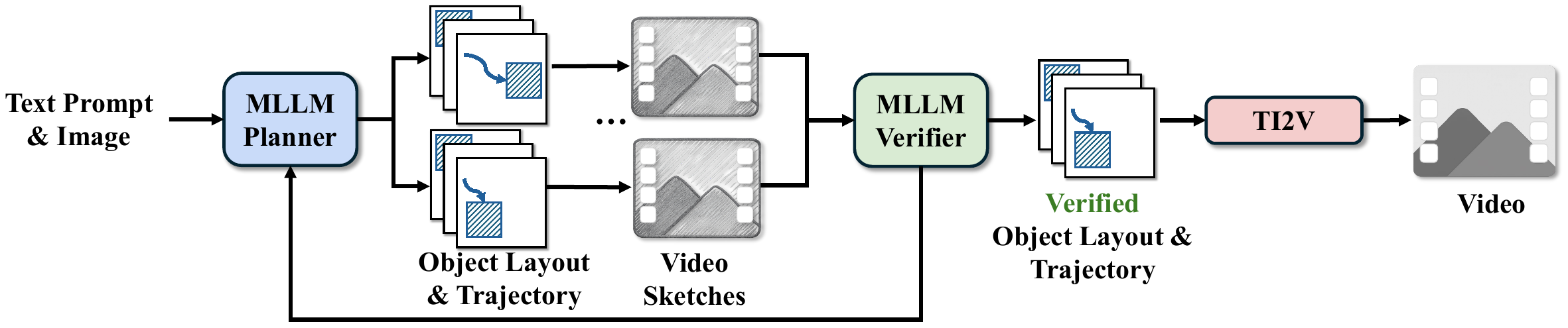}
    \caption{\textbf{\methodname} verifies video sketches before synthesis.}
    \label{fig:teaser-ours}
  \end{subfigure}

  \caption{
 Previous methods trade off correction and generation cost through single-shot planning or iterative refinement. \methodname{}
  selects high-quality plans by verifying lightweight sketches before
  final video synthesis.
  }
  \vspace{-5pt}
  \label{fig:teaser}
\end{figure}

To improve motion controllability and basic physical commonsense, recent work introduces object layouts~\cite{ardino2021click,wang2024levitor} or trajectory plans~\cite{li2025training,huang2025vchain,linvideodirectorgpt} as intermediate controls for video generation, often leveraging large language models (LLMs) or multimodal models for high-level planning. However, most methods rely on single-shot planning (\cref{fig:teaser1}), where one predicted control sequence is directly passed to the generator. Such plans can be noisy or physically inconsistent, with errors propagating to the final video. Iterative refinement methods (\cref{fig:teaser2}) instead update prompts or controls using feedback from generated videos~\cite{xue2025phyt2v,lee2026self,gu2023reuse}, but require repeated time-consuming diffusion calls. This creates a fundamental quality--efficiency trade-off: reliable motion planning benefits from exploring and verifying multiple candidates for instruction alignment and visually observable physical consistency, yet evaluating these candidates through full video synthesis is computationally expensive.

To address this quality--efficiency tension, we propose \textbf{\methodname}, a test-time motion planning framework that scales compute in a lightweight planning space by refining motion plans through verification on video sketches (\cref{fig:teaser-ours}), rather than repeatedly invoking costly full-video generation. \methodname{} integrates a multimodal verifier with test-time search to identify and correct semantic and motion inconsistencies in candidate plans, enabling effective selection among multiple motion hypotheses under constrained test-time budgets. By decoupling plan refinement from the diffusion backbone and performing verification at the sketch level, \methodname{} enables efficient test-time compute scaling while improving instruction following and visually observable physical consistency. Compared with baselines that rely on repeated full-video synthesis, it achieves a 13.1$\times$ planning-stage speedup (\Cref{sec:quantitative}).

Specifically, given a text prompt and an initial image, our method first
decomposes the instruction into sequential sub-instructions (e.g.,
``approach the ball'' and ``pick it up'') and identifies the corresponding
movable objects through segmentation.
For each sub-instruction, \methodname{} samples multiple candidate trajectory
plans, represented as bounding-box sequences over time.
To verify these candidates efficiently, we render each trajectory as a
lightweight \textbf{video sketch} by cropping the segmented object from the
initial frame and compositing it onto a static background along the predicted
path, avoiding full-video synthesis during planning.
A vision--language verifier then scores each sketch along two dimensions:
semantic alignment with the sub-instruction and visual motion coherence across
frames.
The motion-coherence score is obtained through structured reasoning over basic
physical commonsense, including Newtonian consistency, non-penetration,
gravity-coherent motion, and shape stability.
If all candidates score below a threshold, verifier feedback is
used to sample a new batch of trajectories.
The highest-scoring acceptable trajectory is selected for each sub-instruction,
and the selected sub-trajectories are merged into a unified motion plan for
trajectory-conditioned video generation.
By verifying plans before synthesis, \methodname{} produces semantically
coherent videos with visually consistent motion at test time, without
additional training or costly full-video iterative refinement.

We evaluate \methodname{} on \textsc{WorldModelBench} and
\textsc{PhyWorldBench}, two large-scale benchmarks that jointly assess whether
generated videos follow the input instructions and exhibit temporally coherent,
commonsense-consistent motion that visually accords with basic physical expectations.
Under the same ATI-14B generator and
generation settings, \methodname{} improves the aggregate
\textsc{WorldModelBench} score from 7.71 to 8.71, isolating the benefit of the
proposed motion-planning framework. Across broader comparisons, our method
achieves the strongest overall performance on both benchmarks and the best
non-MLLM motion-centric PAS, TCS, and CAS scores, while providing a
13.1$\times$ planning-stage speedup over full-video iterative refinement. Ablations show that multimodal
verification, stronger verifiers, and larger sampling budgets improve trajectory
quality; sketch-based verification matches full-video verification with a
9.5$\times$ efficiency gain; and verifier scores positively correlate with final
video metrics.

\section{Related Works}

\begin{figure*}[t]
    \centering
    \includegraphics[width=0.98\linewidth]{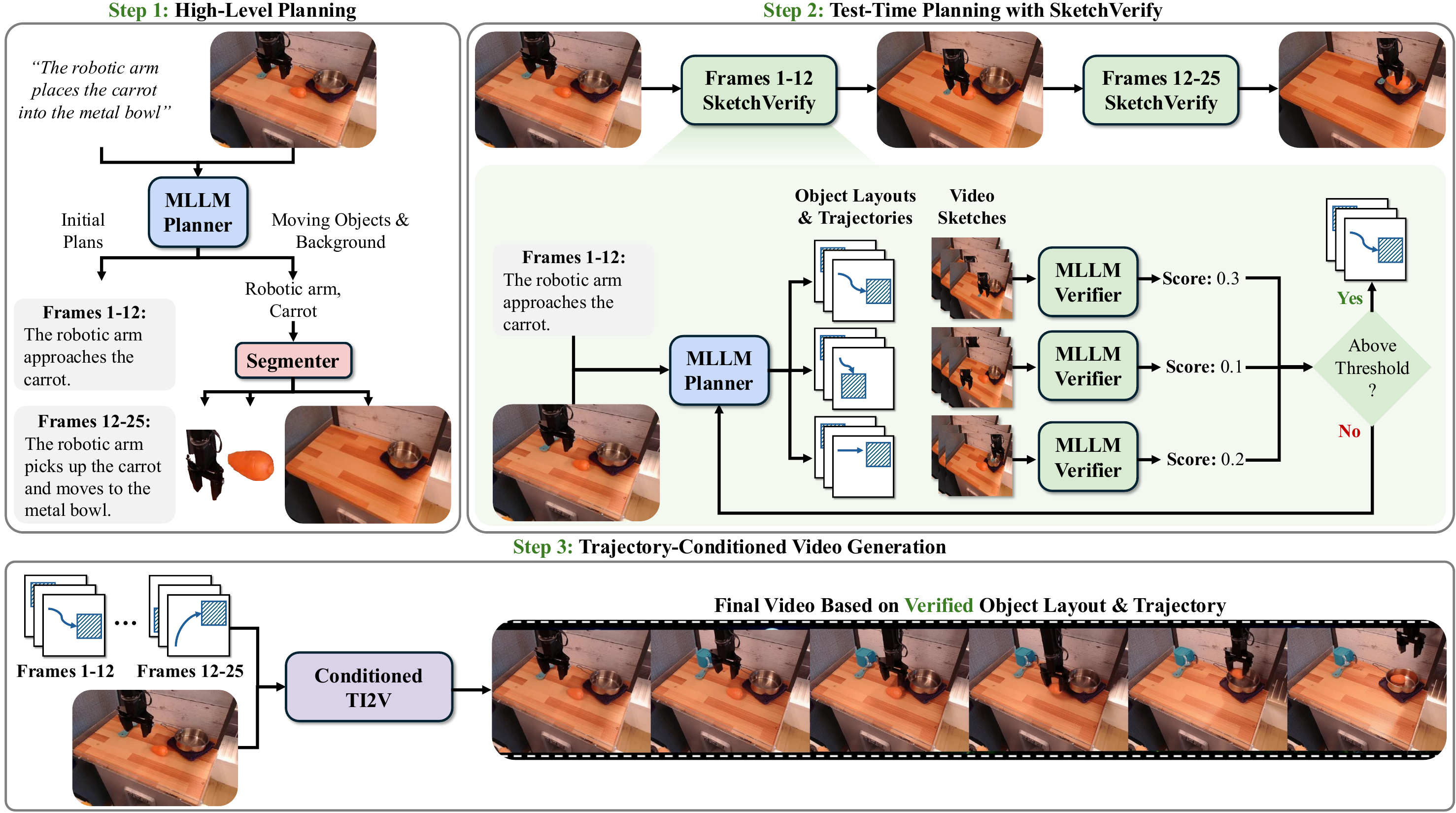}
    \vspace{-8pt}
    \caption{\textbf{Overview of our framework.} 
    Given a prompt and initial frame, we (1) decompose instructions and segment movable objects, (2) sample and verify candidate trajectories using lightweight video sketches scored by a multimodal verifier and (3) synthesize the final video using a trajectory-conditioned diffusion model. }
    \label{fig:pipeline}
\end{figure*}

\label{sec:related_work}

\noindent\textbf{MLLM Planning for Video Generation.}
Recent work increasingly uses LLMs and MLLMs to inject structured planning into video generation, expanding sparse prompts into ``video plans''—bounding-box trajectories~\cite{linvideodirectorgpt, li2025training, wang2024dreamrunner}, scene-level keyframes~\cite{huang2025vchain}, or motion-aware sketches~\cite{li2025training}—that guide layout-guided diffusion synthesis~\cite{linvideodirectorgpt, zhuang2024vlogger, lian2023llm, he2023animate, zhoustorydiffusion, yoon2025raccoon}. However, such single-pass plans are often coarse or physically inconsistent. In contrast, we perform \emph{iterative}, verifier-guided refinement, scoring and updating candidate trajectories for stronger spatial and motion plausibility.

\noindent\textbf{Iterative Refinement for Visual Generation.}
Iterative refinement is widely used for consistency and controllability in visual generation. Methods such as RPG~\cite{yang2024mastering}, PhyT2V~\cite{xue2025phyt2v}, VideoRepair~\cite{lee2026self}, and VISTA~\cite{long2025vista} refine prompts or layouts by re-evaluating fully generated videos, often taking over 30 minutes per cycle. We instead refine at the planning stage, scoring lightweight sketch-based simulations with a multimodal verifier—avoiding repeated video synthesis while preserving verification quality.

\noindent\textbf{Physically Plausible Video Generation.}
State-of-the-art video diffusion models often violate even basic physical laws~\cite{bansal2025videophy,motamed2025generative}. Prior work explores physics-simulator–guided motion~\cite{liu2024physgen, tan2024physmotion, wang2025epic, montanaro2024motioncraft}, physics-driven post-training and reward optimization~\cite{li2025pisa, wang2025physcorr, lin2025reasoning}, and vision–language reasoning or force-based conditioning that injects implicit physical priors~\cite{yang2025vlipp, wang2025physctrl, gillman2025force, hao2025enhancing}. Avoiding heavy simulation, specialized datasets, or finetuning, our SketchVerify performs plan-level iterative refinement, achieving zero-shot generalization across diverse physical scenarios.

\section{Method -- \methodname}
\label{sec:method}

We present \methodname{}, a sketch-guided verification framework for efficient
motion planning in text-and-image-to-video generation. Given a prompt
$\mathcal{P}$ and an initial image $\mathbf{I}_0$, our method produces a
semantically aligned video with temporally coherent motion,
$\mathbf{V}=\{\mathbf{I}_1,\dots,\mathbf{I}_T\}$, by decoupling motion planning
from video synthesis. This enables structured trajectory proposal,
sketch-based verification, ranking, and selection before expensive pixel-level
generation.
The remainder of this section is organized as follows.
\cref{sec:method_parsing} describes plan decomposition and object/background
extraction. \cref{sec:method_vtts} presents our visual test-time planning module,
which samples video sketches and verifies their semantic alignment and physical
plausibility. \cref{sec:final_gen} describes trajectory-conditioned final video
synthesis.

\subsection{High-Level Planning and Object Parsing}
\label{sec:method_parsing}

\noindent\textbf{High-Level Planning.}
We begin by decomposing the natural language prompt $\mathcal{P}$ into a structured sequence of sub-instructions ${\mathcal{P}_1, \ldots, \mathcal{P}_M}$ (e.g., “approach the ball,” “pick up the ball”) using an MLLM, thereby mitigating the difficulty of long-horizon planning.

\noindent\textbf{Object and Background Extraction.}
To enable motion planning over a clean static canvas, we first identify objects involved in motion. Specifically, given the prompt $\mathcal{P}$ and initial frame $\mathbf{I}_0$, we use an MLLM to extract a list of object names expected to move according to the described actions. 
Next, we apply a detector–segmenter pair, GroundedSAM~\cite{ren2024grounded,kirillov2023segment,liu2023grounding}, for mask extraction to localize the mentioned objects. 
This results in a set of object masks $\mathcal{M} = \{m_1, \ldots, m_N\}$ corresponding to the moving entities, where $N$ is the number of moving objects proposed. 
To obtain a clean, static background for compositing, we remove the masked object regions from $\mathbf{I}_0$ and fill them using Omnieraser~\cite{wei2025omnieraser}, a background inpainting model fine-tuned from FLUX~\cite{flux2024}. 
The output is a static background image $\mathbf{B}$, which serves as the canvas for video sketch rendering in subsequent stages (Step~1 in \cref{fig:pipeline}).

\subsection{Test-Time Planning with \methodname }
\label{sec:method_vtts}

To improve trajectory quality without incurring the cost of iterative synthesis, we propose {\methodname}, a sketch-verification guided test-time planning module that verifies object-level motion plans before video generation (\cref{fig:pipeline}).

\begin{figure}[t]
    \centering
    \includegraphics[width=\linewidth]{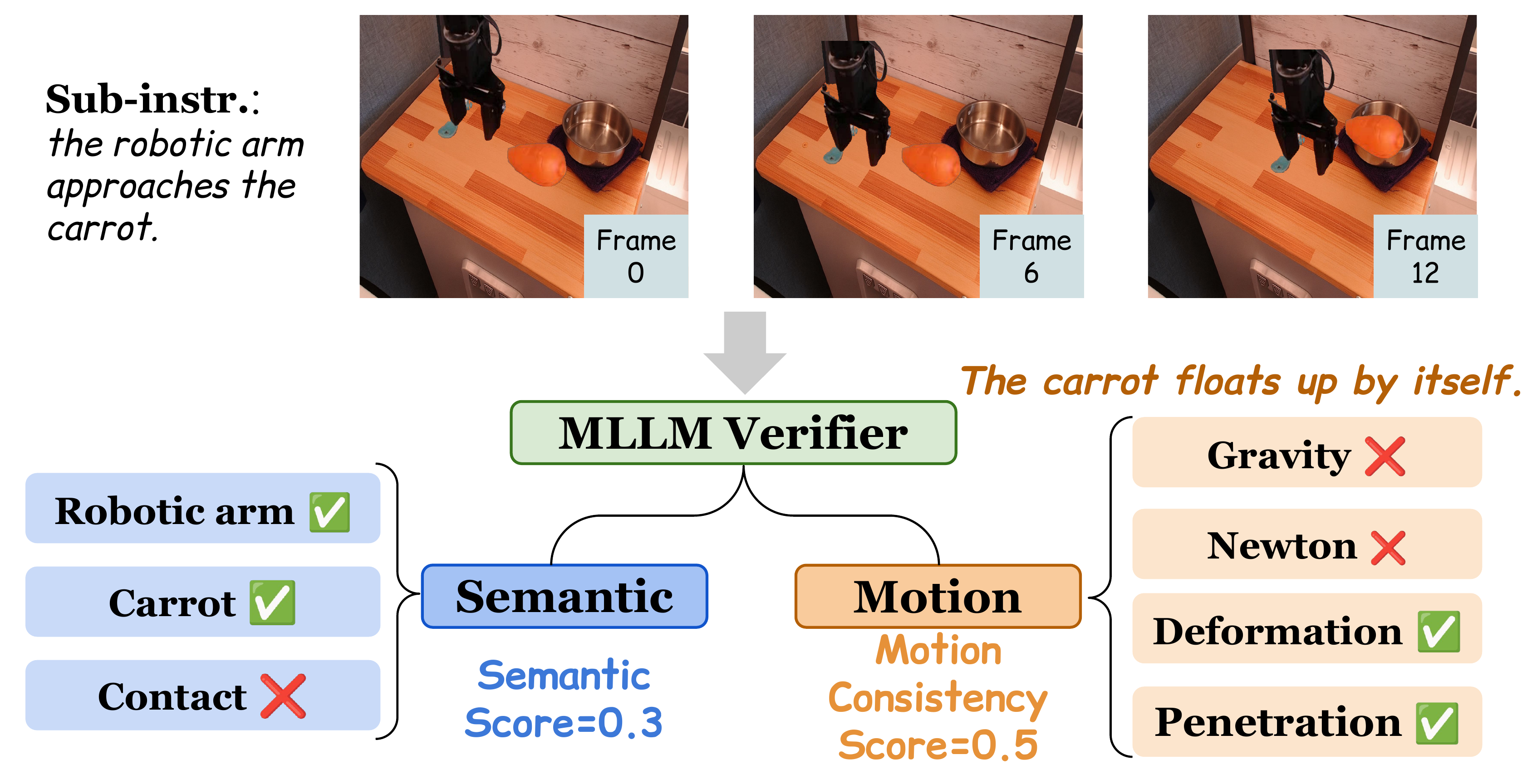}
    \caption{Illustration of MLLM verifier.
    Given a video sketch and sub-instruction, the MLLM outputs semantic-alignment and motion-consistency scores used to rank candidate trajectories.}
    \label{fig:verifier}
    \vspace{-8pt}
\end{figure}
\noindent\textbf{Trajectory Sampling.}
For each sub-instruction $\mathcal{P}_i$, the goal is to generate a set of candidate trajectories that guide the moving object $\mathcal{O}$ according to the intended action. The sampling process is conditioned on the current visual context $\mathbf{C}_i$, which provides spatial grounding for planning.
We use MLLM Planner 
$\mathcal{F}$ to generate $K$ candidate trajectories:$
\left\{ \Pi_i^{(1)}, \dots, \Pi_i^{(K)} \right\} = \mathcal{F}(\mathcal{P}_i, \mathcal{O}, \mathbf{C}_i),$
where each $\Pi_i^{(k)}$ is a list of bounding boxes over $T_i$ frames: $
\Pi_i^{(k)} = \{\, \mathbf{b}_i^{(k,t)} = (x_{\min}^{(t)}, y_{\min}^{(t)}, x_{\max}^{(t)}, y_{\max}^{(t)}) \,\}_{t=1}^{T_i}$
($\mathbf{b}_i^{(k,t)}$ denotes the bounding box of the object at frame $t$, 
with $(x_{\min}^{(t)}, y_{\min}^{(t)})$ and $(x_{\max}^{(t)}, y_{\max}^{(t)})$ being the upper-left and lower-right coordinates, respectively).
For example, if $\mathcal{P}_i$ is “move the apple toward the basket,” then $\Pi_i^{(k)}$ can define a smooth horizontal motion of the apple across $T_i$ frames toward the location of the basket. 
The visual context $\mathbf{C}_i$ is initialized as the reference image $\mathbf{I}_0$ when $i=1$, and updated at each subsequent step to the last frame of the selected sketch $\mathbf{S}_{i-1}^{\ast}$, preserving temporal continuity throughout the planning process.

\noindent\textbf{Video Sketch Rendering.}
To enable assessing plans without incurring the cost of full video generation, we render a lightweight video sketch that visualizes only the intended object motions. Specifically, each trajectory $\Pi_i^{(k)}$ is converted into a sketch $\mathbf{S}_i^{(k)}$ by cropping the segmented object region from the initial frame $\mathbf{I}_0$ using its predicted mask and compositing it frame by frame onto the static background $\mathbf{B}$ according to the bounding box coordinates in $\Pi_i^{(k)}$.
These sketches provide a faithful, layout-preserving approximation of the planned motion, enabling efficient test-time verification focused on spatial–temporal coherence rather than appearance-level generation (see \cref{sec:ablation_verifier}).

\noindent\textbf{Verifier-Guided Scoring.}
As shown in \cref{fig:verifier}, we perform a two-stage evaluation strategy combining semantic alignment and motion consistency using an MLLM. First, we compute a semantic score to evaluate the presence of the target objects and whether the specified action is correctly executed in the proposed motion. Formally, given the sub-instruction $\mathcal{P}_i$ and corresponding sketch $\mathbf{S}_i^{(k)}$, the MLLM $\mathcal{V}_{\text{sem}}$ returns a scalar compatibility score:
$
s_k^{\mathrm{sem}} = \mathcal{V}_{\text{sem}}(\mathbf{S}_i^{(k)}, \mathcal{P}_i)
$.
We evaluate the visual motion consistency of each sketch using few-shot in-context learning across four criteria related to basic physical commonsense:
\textbf{{(i)} Newtonian Consistency:} acceleration and deceleration patterns should align with common-sense physical motion. \textbf{{(ii)} Penetration Violation:} objects should not unrealistically pass through static scene elements. \textbf{{(iii)} Gravitational Coherence:} vertical motion should follow gravity-consistent trends rather than arbitrary displacement. \textbf{{(iv)} Deformation Consistency:} object shape and size should remain visually consistent across frames.
Each response from the MLLM is parsed into a scalar score $s_k^{(l)}$, where $l \in \mathcal{L} =$ \{\text{Newton}, \text{Penetration}, \text{Gravity}, \text{Deformation}\}.
We map descriptive outputs (e.g., ``very consistent'') to numerical values using predefined rules (e.g., ``very consistent'' $\rightarrow$ 1.0, ``somewhat inconsistent'' $\rightarrow$ 0.7).
Detailed prompt templates and mapping rules are provided in Appendix~\ref{sup:prompt}.

The final trajectory is selected by maximizing a weighted combination of semantic-alignment and motion-consistency scores:
$\Pi_i^* = \arg\max_k \left( \lambda_{\text{sem}} s_k^{\text{sem}} + \sum_{l \in \mathcal{L}} \lambda_l s_k^{(l)} \right),$
where $\mathcal{L}$ denotes the set of motion-consistency dimensions and the $\lambda$ coefficients balance their relative importance.

\noindent\textbf{Iterative Trajectory Selection.}
To balance planning quality and efficiency, we adopt a bounded iterative refinement strategy. If all candidate scores fall below a quality threshold $\tau$, the entire set is discarded. A new batch of trajectories is then sampled by prompting the planner with an augmented instruction that incorporates feedback on previous failure cases until a valid plan is found or a retry limit is reached.
Finally, we set the last frame of $\mathbf{S}_i^*$ as the new context frame $\mathbf{C}_{i+1}$ for the next round of planning on the subsequent sub-instruction, preserving temporal continuity across the selected motion plans (Step~2 in \cref{fig:pipeline}).

\subsection{Trajectory-Conditioned Video Generation}
\label{sec:final_gen}
Once the full instruction plan has been verified and selected, we obtain a verified motion plan $\Pi^\ast = \{\Pi_1^\ast, \ldots, \Pi_M^\ast\}$, where each sub-plan $\Pi_i^\ast$ is a sequence of bounding boxes $\{\mathbf{b}_{i,1}, \ldots, \mathbf{b}_{i,T_i}\}$ over $T_i$ frames. For each box $\mathbf{b}_{i,t} \in \mathbb{R}^4$, we extract a representative point $p_{i,t} \in \mathbb{R}^2$ (e.g., its center), resulting in a sparse trajectory $\mathcal{Q}_i = \{p_{i,1},\ldots,p_{i,T_i}\}$.
The full object path is formed by concatenating the sub-trajectories in temporal order:
$\mathcal{Q}^{\ast}
=
\mathcal{Q}_1 \| \ldots \| \mathcal{Q}_M$.
We then temporally interpolate this sequence to obtain a dense trajectory
$\bar{\mathcal{Q}}
=
\{q_1,\ldots,q_T\}$ over $T$ output frames, where each
$q_t \in \mathbb{R}^2$ specifies the object position at time $t$.

We adopt a pre-trained trajectory-conditioned image-to-video diffusion model for final video synthesis. The model encodes the initial image $\mathbf{I}_0$ into latent features and modulates the denoising process by injecting trajectory latents derived from $\bar{\mathcal{Q}}$. These trajectory conditions guide the generation process to produce coherent object motion while preserving the appearance of the initial image and following the intended spatial path. The resulting $T$-frame video $\mathbf{V}$ follows the high-level prompt and is guided by a motion trajectory selected for semantic alignment and visual motion consistency (Step~3 in \cref{fig:pipeline}).
\section{Experimental Results}

\subsection{Experimental Setup}

\noindent\textbf{Benchmarks.}
We evaluate our method on two recent large-scale benchmarks for visual world models: 
\textbf{WorldModelBench}~\cite{li2025worldmodelbench}, which evaluates instruction following, perceived physical plausibility, and commonsense across seven domains using a benchmark-provided MLLM scorer; and \textbf{PhyWorldBench}~\cite{gu2025phyworldbench}, which tests visually observable physical consistency over 350 prompts. Together, these benchmarks capture the two primary outcomes targeted by our motion-planning framework: whether the generated video follows the requested instruction and whether its
motion remains temporally coherent and visually consistent with basic physical
commonsense.
Since PhyWorldBench is a text-to-video benchmark, we generate the first frame using FLUX~\cite{flux2024} and then perform image-to-video generation.

\noindent\textbf{Evaluation Metrics.}
Across the two benchmarks, we evaluate instruction following, benchmark-defined physical-consistency dimensions (Newtonian motion, deformation, fluid dynamics, penetration, and gravity), and commonsense consistency at both the frame and temporal levels. These quality metrics are computed using the benchmark-provided MLLM scorers. We additionally evaluate efficiency in terms of planning time and generation cost. Full metric definitions are provided in Appendix~\ref{sup:evaluation}.

\begin{table*}[t]
\centering
\vspace{-5pt}
\caption{
Comparison on \textsc{WorldModelBench}~\cite{li2025worldmodelbench}.
We report instruction-following, benchmark-defined physical-consistency, and commonsense scores, together with their aggregate \textbf{Sum}.
ATI-14B (generator only) uses the same video generator as \methodname{} without sketch-guided motion planning.
Best results in each column are highlighted in bold.
Planning time is measured in minutes.
}
\vspace{-10pt}
\resizebox{\textwidth}{!}{
\begin{tabular}{lc|ccccc|cc|c|c}
\toprule
\multirow{2}{*}{\textbf{Model}}
& \multicolumn{1}{c|}{\textbf{Instruction}}
& \multicolumn{5}{c|}{\textbf{Physical Consistency}}
& \multicolumn{2}{c|}{\textbf{Commonsense}}
& \multirow{2}{*}{\textbf{Sum$\uparrow$}}
& \multirow{2}{*}{\shortstack{\textbf{Plan Time}\\\textbf{(min)$\downarrow$}}} \\
\cmidrule(lr){2-2}
\cmidrule(lr){3-7}
\cmidrule(lr){8-9}
& \textbf{Follow$\uparrow$}
& \textbf{Newton$\uparrow$}
& \textbf{Deform$\uparrow$}
& \textbf{Fluid$\uparrow$}
& \textbf{Penetr.$\uparrow$}
& \textbf{Gravity$\uparrow$}
& \textbf{Frame$\uparrow$}
& \textbf{Temporal$\uparrow$}
& & \\
\midrule

\rowcolor{gray!15}
\multicolumn{11}{l}{\textbf{Video Generation Models (No Test-Time Planning)}} \\

Hunyuan-Video~\cite{kong2024hunyuanvideo}
& $1.18$ & $\mathbf{1.00}$ & $0.80$ & $\mathbf{1.00}$ & $\mathbf{0.92}$ & $\mathbf{1.00}$
& $0.64$ & $0.70$ & $7.24$ & -- \\

CogVideoX~\cite{yang2024cogvideox}
& $1.46$ & $0.99$ & $0.70$ & $0.99$ & $0.77$ & $0.96$
& $0.86$ & $\mathbf{0.94}$ & $7.67$ & -- \\

Wan-2.1-14B~\cite{wan2025wan}
& $1.88$ & $\mathbf{1.00}$ & $0.76$ & $0.99$ & $0.81$ & $0.99$
& $\mathbf{0.96}$ & $0.82$ & $8.21$ & -- \\

Cosmos~\cite{agarwal2025cosmos}
& $2.06$ & $\mathbf{1.00}$ & $0.84$ & $0.99$ & $\mathbf{0.92}$ & $\mathbf{1.00}$
& $0.92$ & $0.90$ & $8.63$ & -- \\

Open-Sora~\cite{opensora}
& $1.64$ & $0.98$ & $0.82$ & $\mathbf{1.00}$ & $0.91$ & $\mathbf{1.00}$
& $0.80$ & $0.84$ & $7.99$ & -- \\

STEP-Video~\cite{huang2025step}
& $1.04$ & $\mathbf{1.00}$ & $0.75$ & $\mathbf{1.00}$ & $0.89$ & $\mathbf{1.00}$
& $0.50$ & $0.71$ & $6.89$ & -- \\

ATI-14B (generator only)~\cite{wang2025ati}
& $1.92$ & $\mathbf{1.00}$ & $0.67$ & $0.99$ & $0.85$ & $0.98$
& $0.67$ & $0.62$ & $7.71$ & -- \\

\midrule
\rowcolor{gray!15}
\multicolumn{11}{l}{\textbf{Single-Shot Planning}} \\

VideoMSG~\cite{li2025training}
& $1.46$ & $0.99$ & $0.79$ & $0.99$ & $0.83$ & $0.94$
& $\mathbf{0.96}$ & $0.82$ & $7.78$ & $\mathbf{1.33}$ \\

\midrule
\rowcolor{gray!15}
\multicolumn{11}{l}{\textbf{Verifier-Guided Test-Time Planning}} \\

PhyT2V~\cite{xue2025phyt2v}
& $1.97$ & $\mathbf{1.00}$ & $0.82$ & $0.96$ & $0.82$ & $\mathbf{1.00}$
& $0.81$ & $0.81$ & $8.19$ & $61.86$ \\

\textbf{SketchVerify (Ours)}
& $\mathbf{2.08}$ & $\mathbf{1.00}$ & $\mathbf{0.89}$ & $\mathbf{1.00}$ & $\mathbf{0.92}$ & $\mathbf{1.00}$
& $\mathbf{0.96}$ & $0.86$ & $\mathbf{8.71}$ & $4.71$ \\

\bottomrule
\end{tabular}
}
\vspace{-17pt}
\label{tab:worldmodelbench}
\end{table*}
\begin{table}[t]
\caption{
Evaluation on \textsc{PhyWorldBench}. 
We report pass rates for object/event categories, physical standards, and the overall score. 
Best results are in \textbf{bold}; second-best results are \underline{underlined}.
}
\vspace{-5pt}
\centering
\label{tab:phyworldbench}
\small
\setlength{\tabcolsep}{4pt}
\resizebox{\columnwidth}{!}{%
\begin{tabular}{lccc}
\toprule
\textbf{Model} & \textbf{Obj+Evt↑} & \textbf{Phys.~Std↑} & \textbf{All↑} \\
\midrule
CogVideoX~\cite{yang2024cogvideox}       & 41.62 & \underline{21.68} & \underline{17.34} \\
Wan-2.1~\cite{wan2025wan}                    & 35.34 & 19.83 & 15.52 \\
OpenSora~\cite{opensora}                 & 36.86 & 17.43 & 14.00 \\
Cosmos~\cite{agarwal2025cosmos}          & \textbf{48.29} & 15.71 & 14.00 \\
Hunyuan~\cite{kong2024hunyuanvideo}      & 24.86 & 14.16 & 10.12 \\
STEP~\cite{huang2025step}                & 29.51 & 16.33 & 12.89 \\
\midrule
\textbf{SketchVerify}                    & \underline{43.11} & \textbf{23.52} & \textbf{19.84} \\
\bottomrule
\end{tabular}%
}
\vspace{-5pt}
\end{table}

\noindent\textbf{Implementation Details} 
We use the multimodal version of GPT-4.1~\cite{openai2024gpt4_1} as the default planner to generate five candidate trajectories for each sub-instruction, where each trajectory records the coordinates of the top-left and bottom-right corners of the object bounding box at every frame.
We construct lightweight video sketches by pasting foreground objects, segmented with GroundedSAM~\cite{ren2024grounded}, onto static backgrounds to visualize the planned object motion. These sketches are scored in the range $[0,1]$ by Gemini 2.5~\cite{google2025gemini2_5}. This vision--language verifier uses prompt-based queries to assess semantic alignment and visual motion consistency, including Newtonian consistency, object penetration, gravitational coherence, and deformation consistency. We assess each criterion using in-context prompts with positive and negative trajectories and aggregate the resulting scores for final plan ranking.
We use ATI-14B~\cite{wang2025ati} to generate 81-frame videos at 480p resolution. Additional implementation details are provided in Appendix~\ref{sup:otherdeatil}.

\subsection{Quantitative Results}
\label{sec:quantitative}

\noindent\textbf{Evaluation on WorldModelBench.}
\Cref{tab:worldmodelbench} compares our method with strong open-source video generation models, including CogVideoX~\cite{yang2024cogvideox}, Cosmos~\cite{agarwal2025cosmos}, Hunyuan-Video~\cite{kong2024hunyuanvideo}, Open-Sora~\cite{opensora}, Wan-2.1~\cite{wan2025wan}, and STEP-Video~\cite{huang2025step}, as well as planning-based baselines such as { VideoMSG}~\cite{li2025training} (single-shot) and { PhyT2V}~\cite{xue2025phyt2v} (multi-step refinement). 
Our method achieves the highest aggregate score (8.71) and strong performance across instruction following and motion-consistency dimensions. Compared with ATI-14B~\cite{wang2025ati}, \methodname{} improves instruction following from 1.92 to 2.08 and the aggregate score from 7.71 to 8.71. Compared with Wan-2.1-14B~\cite{wan2025wan}, it improves instruction following by $10.6\%$ and increases the deformation pass rate from 0.76 to 0.89.
While multi-step pipelines such as PhyT2V provide gains over single-shot planners, they rely on repeated, computationally expensive synthesis cycles (typically requiring about 70 minutes for full plan-synthesize-refine cycles). In contrast, our verifier-guided sampling performs high-quality trajectory selection before a single final synthesis pass, reducing planning time from 61.86 to 4.71 minutes, corresponding to a $92.4\%$ reduction ($13.1\times$ speedup). We present a detailed per-component runtime analysis in Appendix~\ref{sup:time}.

\noindent\textbf{Evaluation on PhyWorldBench.}
As shown in \Cref{tab:phyworldbench}, our framework achieves the highest overall score (19.84) and the strongest performance in the benchmark-defined physical standard category (23.52), indicating that the generated videos more often exhibit visually consistent motion. While Cosmos~\cite{agarwal2025cosmos} attains a higher object--event score (48.29 vs.\ 43.11), its physical standard and overall scores are substantially lower.
These results highlight that our test-time verification not only preserves object-level realism but also improves temporal coherence and visual motion consistency across diverse scenarios.

\noindent\textbf{Non-MLLM Motion-Centric Evaluation.}
\label{app:non-mllm-eval} To mitigate metric hacking and ensure that improvements are not merely artifacts of the verifier used in our framework, we report motion-centric metrics from VMBench~\cite{ling2025vmbench}, including perceptible amplitude (PAS), temporal coherence (TCS), and commonsense adherence (CAS), computed on videos generated from WorldModelBench prompts. These metrics evaluate motion realism and physical plausibility without relying on MLLM verifiers.
As shown in \Cref{tab:non-mllm-eval}, \methodname{} consistently achieves the best performance across all three metrics. This suggests that the gains observed in our framework are not specific to any particular MLLM verifier, but instead reflect consistent improvements in motion realism and temporal--physical coherence of the generated videos.
\begin{table}[t]
\centering
\vspace{-5pt}
\caption{Non-MLLM physics realism evaluation on WorldModelBench prompts using motion-centric metrics from VMBench~\cite{ling2025vmbench}, including perceptible amplitude (PAS), temporal coherence (TCS), and commonsense adherence (CAS). Higher is better.}
\vspace{-5pt}
\small
\setlength{\tabcolsep}{4pt}
\renewcommand{\arraystretch}{0.9}
\resizebox{\columnwidth}{!}{
\begin{tabular}{lccccccc}
\toprule
\textbf{Metric} &
CogVideoX &
Opensora &
Cosmos &
Wan &
PhyT2V &
VideoMSG &
\textbf{SketchVerify} \\
\midrule
\textbf{PAS}$\uparrow$ &
17.03 &
5.21 &
12.55 &
19.99 &
21.04 &
14.26 &
\textbf{26.73} \\
\textbf{TCS}$\uparrow$ &
94.55 &
96.85 &
91.62 &
96.31 &
92.84 &
94.56 &
\textbf{97.38} \\
\textbf{CAS}$\uparrow$ &
47.91 &
46.38 &
42.91 &
48.44 &
47.13 &
44.26 &
\textbf{50.18} \\
\bottomrule
\end{tabular}
}
\label{tab:non-mllm-eval}
\end{table}

\noindent\textbf{Verifier Score and Output Video Quality.}
To examine whether the verifier scores are predictive of final video quality, we measure the Pearson correlation between the VLM verifier scores and the corresponding final-video evaluation metrics. Specifically, for each video generated from a WorldModelBench prompt, we record the verifier scores associated with the selected sketch and compute their correlations with the final metrics.
As shown in \Cref{tab:verifier_correlation}, the verifier scores are positively correlated with the corresponding final-video metrics. The instruction-alignment verifier score has a correlation of $0.68$ with the final instruction-following metric, while the motion-consistency score has a correlation of $0.62$ with the benchmark physical-consistency metric. These results indicate that sketch-level verifier scores provide a useful signal for ranking candidate motion plans, although final video quality also depends on how faithfully the trajectory-conditioned generator realizes the selected plan.

\begin{figure*}[t]
    \centering
    \includegraphics[width=\linewidth]{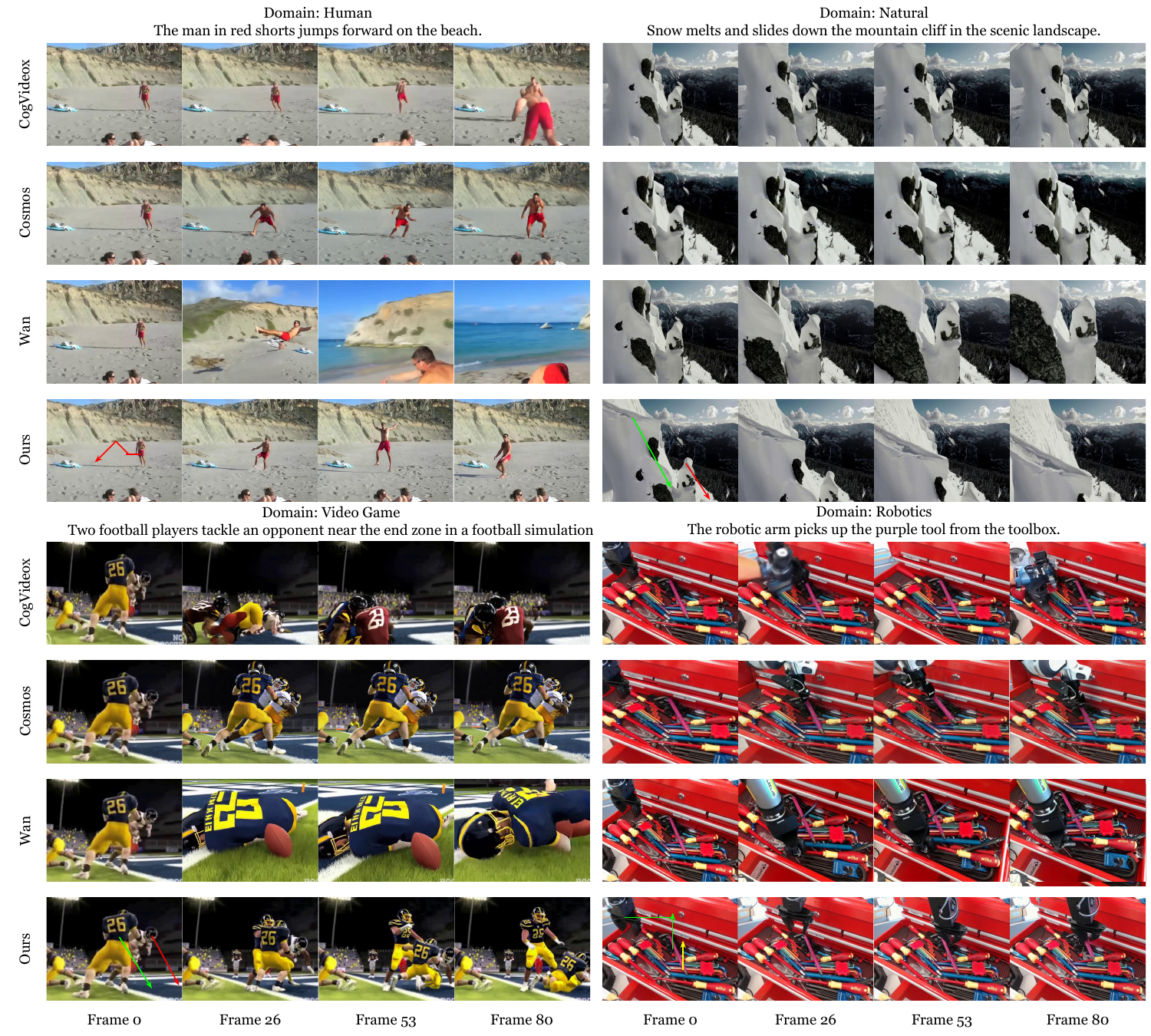}

\vspace{-8pt}
\caption{
Qualitative comparison on four representative domains from \textsc{WorldModelBench}: \textbf{Human}, \textbf{Natural}, \textbf{Video Game}, and \textbf{Robotics}. 
Each group shows sampled frames from competing models given the same text prompt. 
Frames are uniformly sampled from each generated 81-frame video.
}
\vspace{-8pt}
    \label{fig:qualitatve}
\end{figure*}

\subsection{Qualitative Results}
\cref{fig:qualitatve} compares results across four \textsc{WorldModelBench}
domains: Human, Natural, Video Game, and Robotics. Existing baselines such as
CogVideoX~\cite{yang2024cogvideox}, Cosmos~\cite{agarwal2025cosmos}, and
Wan-2.1~\cite{wan2025wan} often exhibit visible physical artifacts. In the Human
domain, baselines produce unnatural body stretching or floating during jumps,
whereas our method generates smoother forward motion with more consistent
gravity response. In the Natural domain, baselines fail to move the snow
according to the instruction, while ours produces continuous downhill motion.
In Video Game and Robotics scenes, competing methods often show collision
misalignment, object merging, or gripper--object mismatch, whereas our method
preserves more accurate contacts and stable manipulation. These examples show
that verifier-guided planning reduces physical implausibilities and improves
temporal coherence across diverse environments. We provide additional samples
in Appendix~\ref{sup:qualitative}.

\begin{table}[t]
\centering
\caption{
Pearson correlation between sketch-level verifier scores and final-video
metrics on WorldModelBench. ``Phys.'' denotes the benchmark-defined
physical-consistency score.
}
\vspace{-5pt}
\small
\setlength{\tabcolsep}{5pt}
\renewcommand{\arraystretch}{0.95}
\begin{tabular}{lcccc}
\toprule
\multirow{2}{*}{\textbf{Verifier Score}}
& \multicolumn{4}{c}{\textbf{Output Video Metric}} \\
\cmidrule(lr){2-5}
& \textbf{Instr.} & \textbf{Phys.} & \textbf{Common.} & \textbf{Sum} \\
\midrule
Instruction alignment & \textbf{0.68} & 0.29 & 0.25 & 0.42 \\
Motion consistency     & 0.27 & \textbf{0.62} & 0.31 & 0.46 \\
\bottomrule
\end{tabular}
\vspace{-5pt}
\label{tab:verifier_correlation}
\end{table}

\subsection{Ablation Study}

We conduct ablation studies on WorldModelBench to examine how verifier type and test-time sampling affect motion planning quality.

{
\noindent\textbf{Effect of Different Planner Choices.}
We compare different MLLMs as planners for candidate trajectory generation (\Cref{tab:planner_strength}). Smaller models such as Qwen2.5-VL-3B yield limited gains, suggesting that weaker spatial reasoning and grounding constrain their ability to produce accurate object trajectories. In contrast, stronger planners such as GPT-4.1 generate more reliable candidate motion plans for subsequent verification. Weaker open-source VLMs also exhibit limited instruction-following ability on multi-step, numerically precise planning tasks, further reducing the quality of the proposed trajectories.

}

\begin{figure*}[t]
    \centering
    \includegraphics[width=\linewidth]{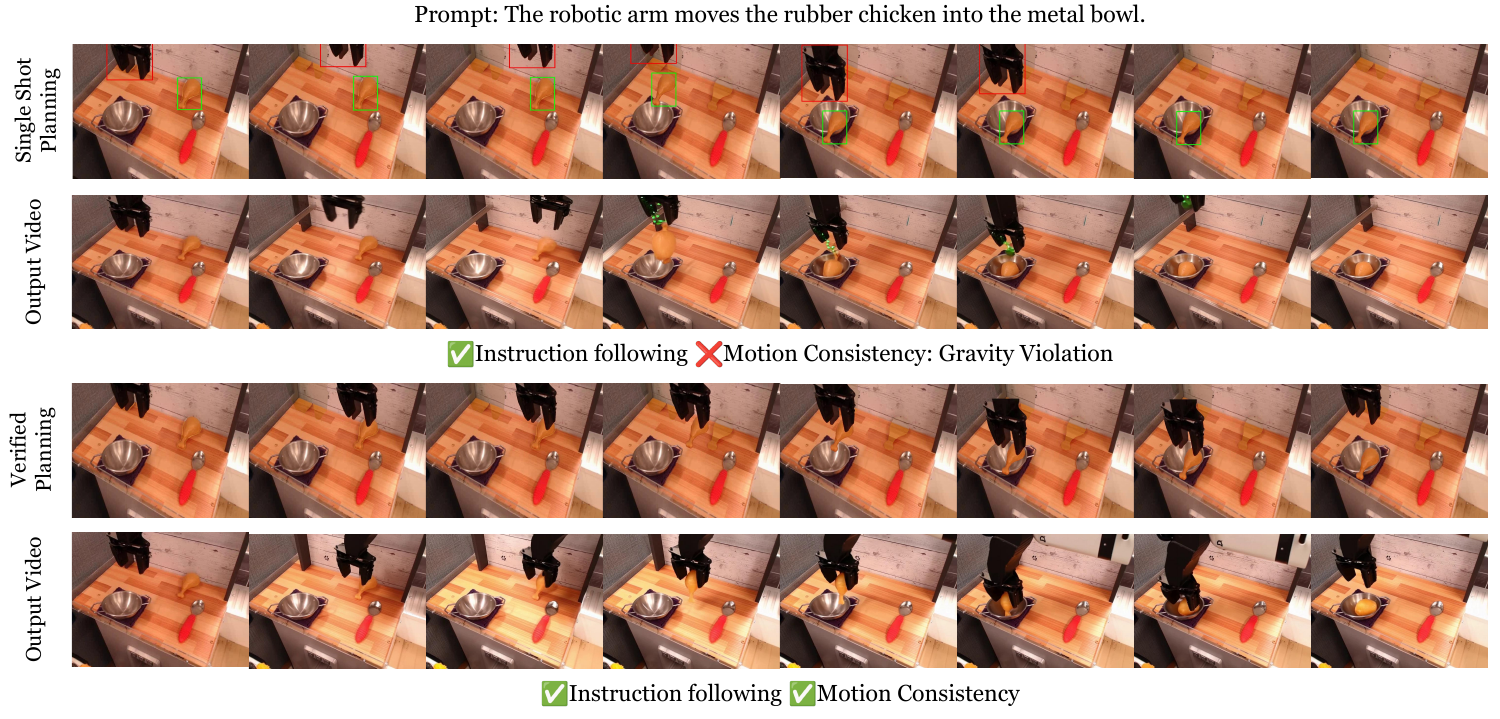}
    \vspace{-20pt}
    \caption{
Ablation study on verifier modality.
Introducing visual input to the verifier significantly improves both instruction following and consistency with basic physical commonsense, highlighting the importance of multimodal grounding for reliable trajectory evaluation.
}
    \label{fig:qualitatve_ablation}
\end{figure*}

\noindent\textbf{Effect of Verifier Choice.}
\label{sec:ablation_verifier}
We first study how verifier modality and scale affect trajectory selection.
As shown in \Cref{tab:ablation_verifier}, language-only verification provides only limited gains over single-shot planning, since textual feedback alone cannot directly capture fine-grained spatial interactions, trajectory smoothness, or visible motion inconsistencies. In contrast, the multimodal verifier substantially improves both instruction following and visual motion consistency by directly assessing the rendered motion before synthesis. As illustrated in \cref{fig:qualitatve_ablation}, verification using video sketches helps identify and reject trajectories containing visible spatial or motion inconsistencies, leading to more coherent final motion.
\Cref{tab:verifier_strength} further shows that stronger MLLM verifiers lead to consistent improvements, suggesting that stronger multimodal reasoning improves plan ranking.

\change{
\noindent\textbf{Verification Strategy.}
We further compare sketch-based verification with full-video verification in \Cref{tab:ablation_sketch}. Due to the high computational cost of full-video generation and verification, we use $K=1$ instead of the default $K=5$ for this comparison. Both methods adopt a two-round verify--regenerate setting, where the planner generates one trajectory per round, receives verifier feedback, and refines the trajectory in the next round. Sketch-based verification achieves comparable performance while reducing planning time from 38.99 to 4.08 minutes, corresponding to a 9.5$\times$ speedup. This result suggests that lightweight sketches preserve sufficient spatiotemporal information for effective verification while avoiding the computational cost of full-video synthesis. We further evaluate longer verify--regenerate budgets in Appendix~\ref{app:verification_rounds}.

}

\begin{table*}[t]
\vspace{-7pt}
\caption{
Ablations on \textsc{WorldModelBench}. We analyze planner scale, verifier modality,
verifier scale, and verification strategy. All methods use the same inference
pipeline; time measures planning and verification before final
diffusion-based generation. (I.: instruction-following score; P.: physical plausibility score; T.: planning time.)
}
\vspace{-7pt}
\centering
\footnotesize
\setlength{\tabcolsep}{3pt}
\renewcommand{\arraystretch}{0.92}
\captionsetup[subtable]{font=footnotesize,skip=1pt}

\begin{subtable}[t]{0.23\textwidth}
\centering
\begin{tabular}{@{}lcc@{}}
\toprule
\textbf{Planner} & \textbf{I.} & \textbf{P.} \\
\midrule
Qwen2.5-3B  & 1.23 & 4.50 \\
Qwen2.5-72B & 1.59 & 4.57 \\
GPT-4.1     & \textbf{2.08} & \textbf{4.81} \\
\bottomrule
\end{tabular}
\subcaption{Planner scale.}
\label{tab:planner_strength}
\end{subtable}
\hfill
\begin{subtable}[t]{0.23\textwidth}
\centering
\begin{tabular}{@{}lcc@{}}
\toprule
\textbf{Verifier} & \textbf{I.} & \textbf{P.} \\
\midrule
Single-shot   & 1.46 & 4.55 \\
Lang-only     & 1.49 & 4.76 \\
\textbf{MLLM} & \textbf{2.08} & \textbf{4.81} \\
\bottomrule
\end{tabular}
\subcaption{Verifier modality.}
\label{tab:ablation_verifier}
\end{subtable}
\hfill
\begin{subtable}[t]{0.23\textwidth}
\centering
\begin{tabular}{@{}lcc@{}}
\toprule
\textbf{Verifier} & \textbf{I.} & \textbf{P.} \\
\midrule
Qwen2.5-3B    & 1.62 & 4.68 \\
Qwen2.5-32B   & 1.83 & 4.72 \\
Qwen3-235B    & 2.02 & 4.77 \\
Gemini        & \textbf{2.08} & \textbf{4.81} \\
\bottomrule
\end{tabular}
\subcaption{Verifier scale.}
\label{tab:verifier_strength}
\end{subtable}
\hfill
\begin{subtable}[t]{0.27\textwidth}
\centering
\begin{tabular}{@{}lccc@{}}
\toprule
\textbf{Strategy} & \textbf{I.} & \textbf{P.} & \textbf{T.} \\
\midrule
Unverified   & 1.52 & 4.56 & 0.00 \\
Generation   & \textbf{1.92} & 4.62 & 38.99 \\
Sketch       & 1.90 & \textbf{4.66} & 4.08 \\
\bottomrule
\end{tabular}
\subcaption{Verification strategy.}
\label{tab:ablation_sketch}
\end{subtable}
\label{tab:ablations_grid}
\vspace{-8pt}
\end{table*}

\begin{figure}[t]
\vspace{-15pt}
    \centering
    \includegraphics[width=\linewidth]{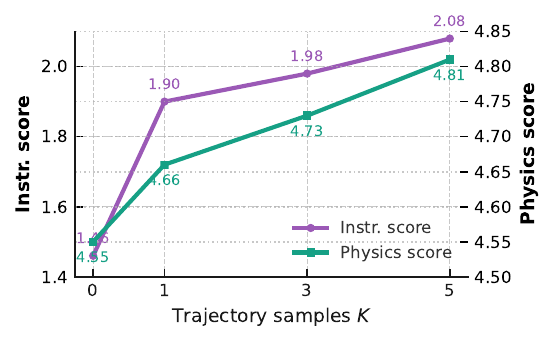}
    \vspace{-20pt}
    \caption{Ablation on the number of sampled trajectories $K$ during planning. Larger $K$ enable stronger verifier-guided selection. $K=0$ denotes the setting without a verifier.
}
\vspace{-10pt}
    \label{fig:ablation_k}
\end{figure}

\noindent\textbf{Effect of Sampling Budget $K$.}~\cref{fig:ablation_k} examines how the number of sampled candidate trajectories $K$ influences the final motion quality. 
When there is no iterative generation (equivalent to {VideoMSG}~\cite{li2025training}), the planner commits to a single trajectory without verification, resulting in limited instruction adherence (1.46) and lower physical consistency (4.55). 
Introducing even a small sampling budget ($K=1$) with refinement yields noticeable gains, and performance continues to increase with larger $K$, reflecting the benefit of exploring a broader trajectory space. 
Our full configuration ($K=5$) achieves the strongest results across both instruction following and physics coherence, showing that moderate test-time sampling is sufficient for robust trajectory selection without compromising efficiency.

\section{Conclusion}
We introduced \methodname{}, a verifier-guided test-time framework that decouples motion planning from video synthesis. By integrating a multimodal verifier into the trajectory-sampling loop and using lightweight video sketches as planning proxies, \methodname{} efficiently evaluates and selects candidate motions before expensive video generation. This design improves instruction following and temporal motion coherence while reducing visible motion artifacts, achieving a 13.1$\times$ planning-stage speedup and a 3.6$\times$ end-to-end speedup over full-video iterative refinement. Experiments on \textsc{WorldModelBench} and \textsc{PhyWorldBench},
together with non-MLLM motion-centric evaluation, demonstrate consistent improvements in instruction alignment, visually observable physical consistency, and overall motion quality. Looking ahead, we aim to extend the framework from 2D image-plane trajectories to 3D scene-aware motion representations, enabling planning and verification in metric world coordinates. Such an extension could support more complex reasoning over depth, occlusion, contact, camera motion, and multi-object interactions, broadening the applicability of sketch-guided motion planning to video world models and embodied simulation.

\section{Acknowledgements}
We thank Justin Chih-Yao Chen, Jaemin Cho, Elias Stengel-Eskin, Zaid Khan, and Shoubin Yu for their helpful feedback.
This work was supported by NSF-AI Engage Institute DRL-2112635, ARO Award W911NF2110220, ONR Grant N00014-23-1-2356, DARPA ECOLE Program No. HR00112390060, and a Capital One Research Award. The views contained in this article are those of the authors and not of the funding agency.
{
    \small
    \bibliographystyle{ieeenat_fullname}
    \bibliography{ref}
}

\clearpage
\appendix
\setcounter{page}{1}

\makeatletter
\renewcommand*{\theHsection}{app.\Alph{section}}
\renewcommand*{\theHsubsection}{app.\Alph{section}.\arabic{subsection}}
\makeatother

\section*{Appendix}
\section{Prompt Templates for the \methodname{} Pipeline}
\label{sup:prompt}
We provide the full set of prompt templates used in our system, covering high-level planning in~\cref{prompt:high-level}, object proposal in~\cref{prompt:object-proposal}, trajectory generation in~\cref{prompt:trajectory-proposal}, plan-alignment verification in~\cref{prompt:alignment-verify}, and motion-consistency verification in~\cref{prompt:physics-verify}. These prompts define the behavior of each module and ensure consistent outputs across tasks and benchmarks. The motion-consistency verifier operationalizes basic physical commonsense through four visually observable criteria---Newtonian consistency, non-penetration, gravitational coherence, and scale stability---rather than claiming formal physical correctness. For reproducibility and future research, we include both system and user instructions exactly as used in our experiments.
\section{Detailed Implementations}
\label{sup:otherdeatil}

\subsection{SketchVerify Pipeline Specification}
\label{sup:implementation}
In this section, we detail the full \methodname{} implementation pipeline, including high-level planning, object detection and masking, background extraction, test-time trajectory search, sketch rendering, multimodal verification, and final video generation. 

\noindent\textbf{High-Level Planning.}
Given the input text prompt, we first perform a high-level decomposition of the described action into a sequence of structured sub-instructions.
This step is carried out by GPT-4.1 (multimodal version) using a constrained prompt that requires the model to output:
(a) a list of action segments,
(b) their temporal ordering, and
(c) the moving objects involved in each segment (detailed prompt in ~\cref{prompt:high-level}).
The planner is asked to produce $M$ sub-instructions ($M \in [1,4]$), depending on the complexity of the prompt.
Each sub-instruction corresponds to an independent phase of motion planning and receives its own temporal budget $T_i$, with all phase lengths summing to the fixed $\sum_{i=1}^M T_i = 41$ total frames used throughout the pipeline.
All high-level plans are parsed through a strict JSON schema that enforces the required fields (action, duration, object ids).
Malformed or under-specified outputs are automatically rejected, and the planner is re-sampled.
This guarantees that downstream modules always operate on well-structured, machine-readable action plans.

\noindent\textbf{Object Detection and Masks.}
We use Grounding~DINO~\cite{ren2024grounded,liu2023grounding} for text-conditioned object detection (based on the detected objects in high-level planning) with a confidence threshold of $0.3$, followed by SAM-HQ~\cite{kirillov2023segment} for segmentation.
For each detected moving object, we retain the highest-scoring instance mask and compute the corresponding bounding box.
Boxes are normalized to $[0,1]^2$ following the image coordinate system used throughout the planner.

\noindent\textbf{Background Removal.}
To obtain a clean static background, we use FLUX.1-dev~\cite{flux2024} with the Omnieraser~\cite{wei2025omnieraser} LoRA.
All moving-object masks are combined into a single inpainting mask.
The background is generated at the same resolution as the input image.
We generate the background image with 28 diffusion steps and $cfg = 3.5$.

\noindent\textbf{Test-Time Search.}
We adopt GPT-4.1 as the default multimodal planner.
Each planning call produces $K = 5$ candidate trajectories, each of length $T_i$, where each trajectory contains per-frame bounding box coordinates for all moving objects.
All planner outputs are validated through a structure-enforcing JSON parser, and malformed samples are automatically re-sampled.
We set the temperature to $1.0$.
Diversity filtering is enforced by requiring an $\ell_2$ distance of at least $0.05$ (in normalized coordinates) between trajectories.

\noindent\textbf{Sketch Rendering.}
For each candidate trajectory, we generate a lightweight {video sketch} of $T_i$ frames using object sprites cropped from the first frame $I_0$ and composited onto the static background.
All sketches are rendered at the input image resolution and saved as MP4 at 4\,fps for verification.

\noindent\textbf{Multimodal Verification.}
We use Gemini~2.5-Flash as the default verifier.
Two scores are produced for each candidate:
\begin{itemize}
    \item a {semantic alignment score} from the first/last-frame comparison;
    \item a {motion-consistency score} from the full sketch video.
\end{itemize}

For scoring, we use the following criteria:
\begin{itemize}
    \item {1.0:} Perfect plan alignment and visually coherent motion.
    \item {0.7--0.9:} Good alignment with minor deviations.
    \item {0.4--0.6:} Partial alignment with some correct aspects.
    \item {0.0--0.3:} Poor alignment; does not achieve the goal or contains clear motion inconsistencies, such as teleportation, penetration, unsupported floating, or unstable object scale.
\end{itemize}

\noindent\textbf{Verifier Weights and Acceptance Threshold.}
In all experiments, we set the semantic-alignment weight to $\lambda_{\mathrm{sem}}=0.5$ and assign an equal weight of $\lambda_l=0.125$ to each of the four motion-consistency criteria, such that the combined weights sum to $1$. We set the acceptance threshold to $\tau=0.7$. The highest-scoring trajectory is accepted if its combined verifier score reaches this threshold; otherwise, the candidate set is discarded and a new batch is sampled using feedback from the rejected candidates. We allow at most three retries for each sub-instruction. If no candidate reaches $\tau$ after the third retry, we select the trajectory with the highest combined verifier score among all candidates sampled across the initial and retry rounds.

\noindent\textbf{Video Generation Model.}
The sub-trajectory lengths sum to 41 planning frames. After trajectory
selection, we temporally interpolate the concatenated trajectory to match
the 81-frame output of ATI-14B. We use the ATI-14B model~\cite{wang2025ati} to generate 81-frame 480p videos.
We generate with 40 steps and $cfg = 5.0$.
The conditions include the input image, the input text prompt, and the trajectory plan obtained from the planner.

\subsection{Baseline and Hardware Specifications}
\label{sup:baseline}
We use Wan2.1-14B-480p-I2V~\cite{wan2025wan}, CogVideoX-5B~\cite{yang2024cogvideox}, Cosmos-Predict2-2B~\cite{agarwal2025cosmos}, HunyuanVideo-I2V~\cite{kong2024hunyuanvideo}, OpenSora-I2V~\cite{opensora}, and Step-Video-I2V~\cite{huang2025step} as baseline open-source TI2V models.
For these models, we directly use the official implementations and sample videos at 480p with 81 frames using 50 diffusion steps.
For PhyT2V and VideoMSG, we replace the backbone video generation model with Wan2.1 for fair comparison.
All experiments are conducted on NVIDIA A100 80G and NVIDIA RTX A6000 GPUs.
\subsection{Per-step Runtime}
\label{sup:time}
On average, high-level planning takes 14.16\,s.
Object detection, segmentation, and background inpainting require 108\,s.
For each sub-instruction, test-time planning takes 72.5\,s on average, consisting of 20.3\,s for trajectory sampling and 52.2\,s for multimodal verification.
All timings are measured on a single NVIDIA A100 GPU using the standard-speed APIs of GPT-4.1 and Gemini-2.5.

\subsection{Cost, Efficiency, and Practicality}
\label{app:cost-efficiency}

We report a per-video cost and runtime breakdown of the SketchVerify pipeline. On average, SketchVerify uses 10.2 planner calls (approximately \$0.15) and 20.9 verifier calls (approximately \$0.13). 

Compared to PhyT2V (70 minutes per video under comparable settings), SketchVerify achieves over $3.6\times$ faster end-to-end generation. This result demonstrates the practical advantage of verifying and refining motion plans in the lightweight sketch space before invoking the video generator.

\section{Experiment Addendum}
\label{sup:evaluation}

\subsection{Benchmark Details}

\noindent\textbf{WorldModelBench}~\cite{li2025worldmodelbench} evaluates video generation models as world models across instruction following, physical plausibility, and commonsense temporal behavior. These dimensions jointly assess properties central to motion planning: whether the generated video realizes the requested action, maintains coherent motion over time, and remains visually consistent with basic physical commonsense.
It contains 7 domains and 56 subdomains across 350 image/text-conditioned tasks.
It supports both I2V and T2V settings; we use I2V.

\noindent\textbf{PhyWorldBench}~\cite{gu2025phyworldbench} evaluates visually observable physical consistency in generated videos, including Newtonian behavior, gravitational motion, and object--event consistency. Its criteria complement instruction-following evaluation by testing whether the motion depicted in the generated video develops coherently and accords with basic physical commonsense. The benchmark contains 350 underlying scenarios, each described using three levels of prompt detail---Event, Physics-Enhanced, and Detailed Narrative---yielding 1,050 prompts in total. We use the most detailed prompt variant for each scenario, resulting in 350 evaluation prompts. Since PhyWorldBench is a text-to-video benchmark, we generate the first frame using FLUX~\cite{flux2024} and then perform image-to-video generation.

\subsection{Metric Details}
\label{sup:metric_detail}
\noindent\textbf{WorldModelBench} provides a vision–language model (MLLM) scorer trained on 67K human annotations.
Each generated video outputs scalar scores in three categories:
\begin{itemize}
    \item {Instruction Following:}  
    Measures how well the generated motion follows the input instruction.  
    Scores range from $0$ to $3$: $3$ indicates that the instructed action is
completed, $2$ indicates partial completion, $1$ indicates motion that does
not follow the intended action, and $0$ indicates that the target subject is
absent or remains stationary.

    \item {Physics Coherence:}
    This benchmark-defined category evaluates visually observable consistency with natural physical priors across five dimensions, each in $[0,1]$: Newtonian motion, deformation consistency, fluid dynamics, object penetration and gravity coherence.

    \item {Commonsense Consistency:}  
    Includes per-frame visual realism and motion smoothness/continuity, both in $[0,1]$.
\end{itemize}

All metrics are computed by the MLLM via prompt-driven scoring heads.

\noindent\textbf{PhyWorldBench} uses its own MLLM-based evaluator and reports three pass-rate metrics.
For each video, eight frames are sampled uniformly and passed to a proprietary state-of-the-art MLLM.
Questions about the following criteria are asked:
\begin{itemize}
    \item {Obj+Evt:} whether the described objects appear and the event occurs.
    \item {Phys.~Std:} the benchmark-defined physical-standard criterion, which checks whether the depicted motion is consistent with gravity, collision response, continuity, and non-penetration.
    \item {All:} counted as correct only if both Obj+Evt and Phys.~Std pass.
\end{itemize}
Each is computed as a binary decision per prompt and averaged into a percentage.
We use GPT-5 as the proprietary MLLM evaluator.

\subsection{Robustness and Error Analysis}
\label{app:robustness-error}

We conduct a manual error analysis on 100 randomly sampled videos to analyze failure modes in the SketchVerify pipeline. We identify the following primary sources of error:

\begin{itemize}
    \item \textbf{Incorrect object segmentation (8\%):} Imperfect foreground/background separation leads to flawed sketch inputs.
    \item \textbf{Inpainting artifacts (1\%):} Occasional inpainting failures introduce visual inconsistencies.
    \item \textbf{Verifier false positives (14\%):} The verifier may select trajectories that violate the instruction or exhibit visibly inconsistent motion.
    \item \textbf{Plan--generation mismatch (21\%):} Valid plans may degrade during final video synthesis.
    \item \textbf{Candidate exhaustion (32\%):} In some cases, all candidate trajectories are rejected by verification.
\end{itemize}

These results highlight both the robustness of verification-driven motion-plan search and the remaining bottlenecks introduced by segmentation quality and the gap between an abstract trajectory plan and its realization by the video generator.

\begin{table}[t]
\centering
\small
\setlength{\tabcolsep}{4pt}
\renewcommand{\arraystretch}{0.95}
\begin{tabular}{lcccc}
\toprule
\textbf{Strategy} & \textbf{Rounds} & \textbf{I.} & \textbf{P.} & \textbf{T.} \\
\midrule
Generation & 2 & 1.92 & 4.62 & 38.99 \\
Sketch     & 2 & 1.90 & 4.66 & 4.08  \\
\midrule
Generation & 4 & 1.97 & 4.70 & 79.46 \\
Sketch     & 4 & 1.93 & 4.67 & 8.37  \\
\midrule
Generation & 8 & 1.96 & 4.72 & 157.83 \\
Sketch     & 8 & 1.94 & 4.71 & 16.91 \\
\bottomrule
\end{tabular}
\caption{
Effect of verification rounds on \textsc{WorldModelBench}. We compare
generation-based and sketch-based verification under different
verify--regenerate budgets. I.: instruction-following score; P.: physical
plausibility score; T.: approximate planning time by minute.
}
\label{tab:verification_rounds}
\end{table}
\subsection{Additional Analysis on Verification Rounds}
\label{app:verification_rounds}

To further examine whether sketch-based verification remains a reliable proxy
under larger refinement budgets, we compare sketch-based and generation-based
verification with different numbers of verify--regenerate rounds. This analysis
addresses the concern that the two-round comparison in the main paper may
underestimate the benefit of full video verification, since additional rounds
could allow generation-based feedback to better capture appearance changes that
are not visible in lightweight sketches.

For both settings, we use the same planner, verifier, and trajectory-conditioned
generator. In each round, the planner proposes one candidate trajectory, the
verifier evaluates the rendered candidate, and the planner receives verifier
feedback to refine the next trajectory. The only difference is the verification
medium: generation-based verification evaluates fully synthesized videos, while
sketch-based verification evaluates lightweight video sketches before synthesis.
We report the instruction-following score, the benchmark's physical-plausibility
score, and planning time on \textsc{WorldModelBench}.

As shown in \Cref{tab:verification_rounds}, increasing the number of refinement
rounds improves both sketch-based and generation-based verification. However,
generation-based verification incurs substantially higher cost as the number of
rounds increases, while sketch-based verification maintains much lower planning
time. The gap between sketch and generation verification remains small across
larger budgets, suggesting that sketches preserve the key spatial--temporal
information needed for trajectory selection. At the same time, this comparison
also clarifies the limitation of sketch verification: because sketches use
static object composites, they may not capture appearance changes, deformation,
or fine-grained visual artifacts that emerge only during full video synthesis.
\section{Extra Qualitative Results}
\label{sup:qualitative}

\begin{figure*}[t]
    \centering
    \includegraphics[page=1,width=\linewidth]{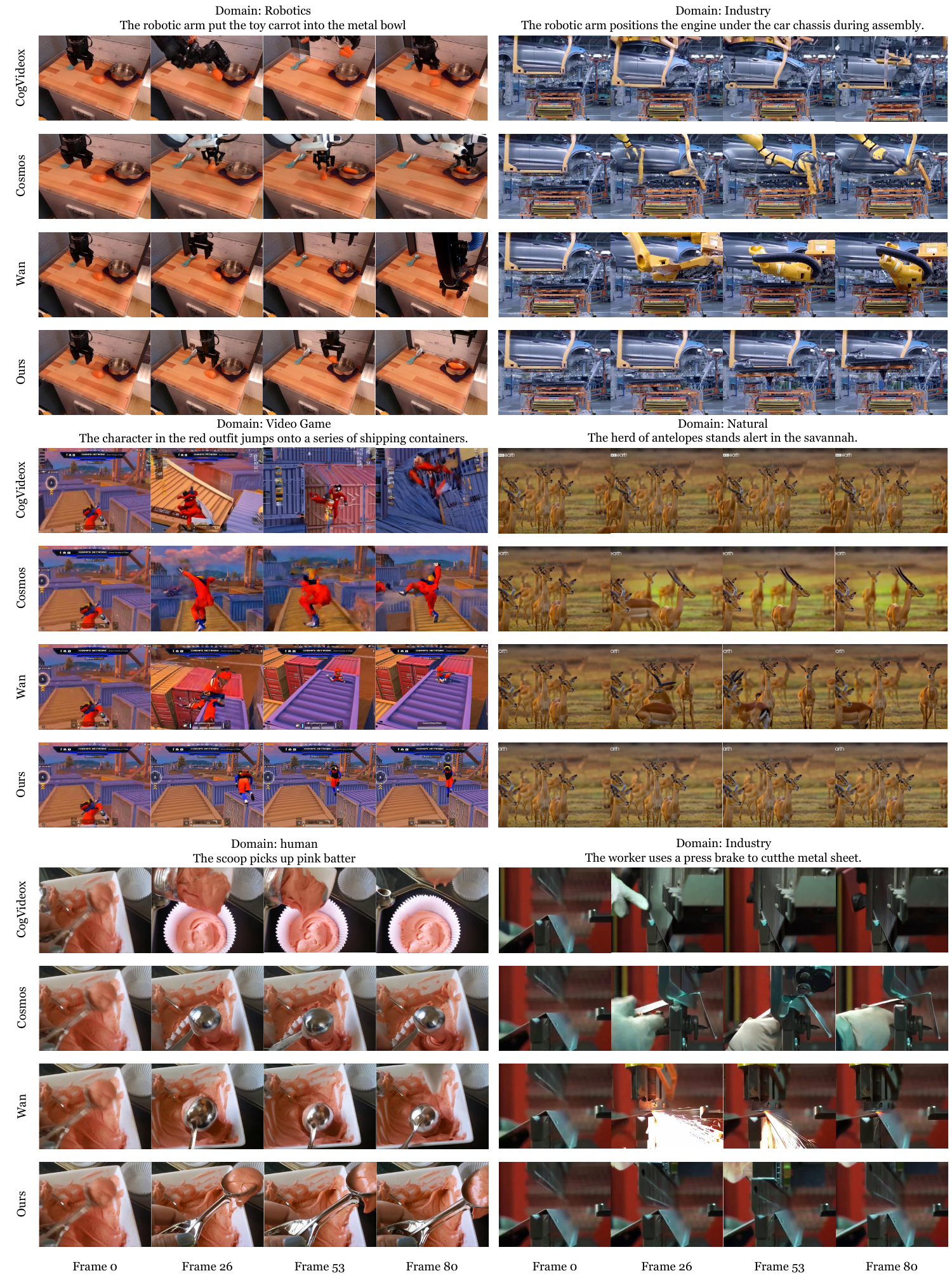}
    \caption{Qualitative comparison on WorldModelBench, highlighting instruction following and coherent motion in the generated videos.}
    \label{fig:extra1}
\end{figure*}

We provide additional qualitative examples in~\cref{fig:extra1} and~\cref{fig:extra2}. These examples illustrate the effectiveness of \methodname{} across diverse scenes and motion types. On WorldModelBench tasks (\cref{fig:extra1}), sketch-guided planning produces generated videos with smoother motion and better alignment with the requested action than the baselines. On PhyWorldBench (\cref{fig:extra2}), our method produces motion that is more visually consistent with basic physical commonsense, avoiding visible artifacts such as unsupported floating and implausible self-motion.

\begin{table}[t]
\centering
\small
\begin{tabular}{p{0.48\linewidth} p{0.42\linewidth}}
\toprule
\textbf{Artifact} & \textbf{License / Terms} \\
\midrule
WorldModelBench~\citep{li2025worldmodelbench} & CC BY 4.0 \\
PhyWorldBench~\citep{gu2025phyworldbench} & MIT License \\
HunyuanVideo~\citep{kong2024hunyuanvideo} & Tencent Hunyuan Community License Agreement \\
CogVideoX~\citep{yang2024cogvideox} & Apache-2.0 \\
Wan2.1~\citep{wan2025wan} & Apache-2.0 \\
Cosmos~\citep{agarwal2025cosmos} & Apache-2.0 \\
Open-Sora~\citep{opensora} & Apache-2.0 \\
Step-Video~\citep{huang2025step} & MIT License \\
VideoMSG~\citep{li2025training} & CC BY 4.0 \\
PhyT2V~\citep{xue2025phyt2v} & CC BY-NC-SA 4.0 \\
ATI~\citep{wang2025ati} & Apache-2.0 \\
\bottomrule
\end{tabular}
\caption{
Summary of scientific artifacts used in this work. We cite the creators of all
artifacts and use them only for academic research purposes consistent with their
licenses or terms of use.
}
\label{tab:artifacts}
\end{table}
\section{License}
\Cref{tab:artifacts} summarizes the scientific artifacts used in this work,
including video generation models, evaluation benchmarks, and datasets. We cite
the creators of all artifacts and list their licenses or terms of use. All
artifacts are used only for academic research and evaluation purposes, consistent
with their intended use and applicable license conditions.

\section{LLM Usage}
We used large language model (LLM) assistants for code debugging and for improving the clarity, wording, and grammar of the manuscript. All technical content, experimental design, analysis, claims, and final writing were reviewed and verified by the authors. The LLMs were not used to generate new scientific claims, conduct experiments autonomously, or make decisions about the reported results.

\begin{figure*}
    \centering
    \includegraphics[page=2,width=\linewidth]{imgs/Supp_Qualitative.pdf}
    \caption{Qualitative comparison on PhyWorldBench, highlighting visually plausible and temporally coherent motion in the generated videos.}
\label{fig:extra2}
\end{figure*}

\begin{figure*}
\begin{tcolorbox}[title=Global Movement Planning Prompt (GPT-4.1 with Tool Calling), colback=white]
\scriptsize

\textbf{System message:}\\
\texttt{You are a video motion planning expert. You have EXACTLY \{total\_frames\} frames to complete the ENTIRE task.}

\textbf{User message:}\\
\texttt{
Text prompt: "\{text\_prompt\}"\\
Available frames: EXACTLY \{total\_frames\}.\\
Current objects in frame 1:\\
- \{label\_1\}: currently at [x\_min, y\_min, x\_max, y\_max]\\
- \{label\_2\}: currently at [...]\ (etc.)\\
\\
Return ONLY a call to \texttt{submit\_movement\_plan} with a complete plan that finishes by frame \{total\_frames\}.}

\textbf{Function schema (\texttt{submit\_movement\_plan}):}
\begin{verbatim}
{
  "task_breakdown": {
    "complete_objective",
    "phase_1", "phase_2", "phase_3",...
    "success_criteria"
  },

  "frame_allocation": {
    "phase_1",
    "phase_2",
    "phase_3",...
  },

  "moving_objects": [...],
  "static_objects": [...],

  "detailed_timeline": {
    frame_1, frame_3, frame_6, frame_8,
    frame_10, frame_12, frame_15, frame_18,...
    frame_{total_frames}
  },

  "movement_plans": {
    "<object_name>": {
      "movement_type",
      "total_distance",
      "movement_phases": {
        "phase_1_frames",
        "phase_2_frames",
        "phase_3_frames",...
      },
    }
  },

  "completion_verification"
}
\end{verbatim}

\end{tcolorbox}

\caption{Prompt used for high-level planning}
\label{prompt:high-level}
\end{figure*}

\begin{figure*}

\scriptsize
\begin{tcolorbox}[
    colback=gray!5,
    colframe=black,
    title={Object Proposal Prompt (System + User)},
    fonttitle=\bfseries,
    arc=2mm
]
\scriptsize
\textbf{System Message:}
\begin{verbatim}
You are an expert in video generation and object-centric scene analysis.
Given the first frame of a video and a text description, determine which
objects should be added, moved, or animated to fulfill the described action.

Your responsibilities:
1. Analyze the given frame and identify all existing objects.
2. Based on the text prompt, determine which additional objects (if any)
   must be introduced to achieve the described event.
3. Identify which objects—either existing or newly added—must move or
   animate to satisfy the prompt.
4. Ensure that proposed object placement and motion are physically
   plausible and consistent with real-world interactions.
5. Focus on major objects that materially affect the scene. If multiple
   parts form a single rigid object, treat them as one entity.
6. For each object name, use minimal wording (1–3 words), concise and
   unambiguous.

Return the result strictly in the following JSON format:

{
  "scene_analysis": "Brief description of the current frame",
  "existing_objects": ["..."],
  "objects_to_add": [
    {
      "name": "object_name",
      "reasoning": "why this object is required",
      "movement_type": "static / linear / curved / complex",
      "priority": "high / medium / low"
    }
  ],
  "moving_objects": ["..."],
  "static_objects": ["..."]
}
\end{verbatim}

\textbf{User Message:}
\begin{verbatim}
Text prompt: "<TEXT_PROMPT>"

Using the first-frame image provided, analyze how the scene should be
modified or animated to satisfy the text description.

Please determine:
- Which required objects are currently missing, based on the prompt.
- Which objects must move or animate to create the described action.
- Which objects remain static as part of the background.
- Realistic object placement and timing relative to the prompt's intent.

Produce the full JSON output exactly as specified in the system instructions.
\end{verbatim}
\end{tcolorbox}
\caption{Prompt used for object proposal}
\label{prompt:object-proposal}\end{figure*}

\begin{figure*}\begin{tcolorbox}[
    colback=gray!5,
    colframe=black,
    title={Sub-Instruction Trajectory Planning Prompt (GPT-4.1)},
    fonttitle=\bfseries,
    arc=2mm
]
\scriptsize
\textbf{System Message:}
\begin{verbatim}
You are a video motion planning expert generating trajectories for frames
{CHUNK_START} to {CHUNK_END}.

Current phase: {PHASE_NAME}
Phase description: {PHASE_DESCRIPTION}
Total frames in video: {TOTAL_FRAMES_NUM}

**COORDINATE SYSTEM:**
{COORDS_GUIDE}

**DIRECTIONAL MAPPINGS:**
- RIGHT: x1 += delta; x2 += delta
- LEFT:  x1 -= delta; x2 -= delta
- UP:    y1 -= delta; y2 -= delta
- DOWN:  y1 += delta; y2 += delta

Focus ONLY on moving objects: {MOVING_OBJECTS}
Ignore static objects in outputs: {STATIC_OBJECTS}
\end{verbatim}

\textbf{User Message:}
\begin{verbatim}
Text prompt: "{TEXT_PROMPT}"

{HISTORY_TEXT}

Generate a smooth trajectory from frame {CHUNK_START} to frame {CHUNK_END}
for this {PHASE_NAME}.

IMPORTANT: Since multiple trajectories will be generated, explore DIFFERENT
valid motion paths. Consider variations in:
- Path shape (straight, curved, arc)
- Speed profile (constant, accelerating, decelerating)
- Intermediate waypoints (different approaches to the goal)

For each object, you should maintain its size (box dimensions) and only
change its position unless you are specifically instructed to resize.

For EACH frame from {CHUNK_START} to {CHUNK_END}, output:
Frame_N: [["object_name", [x1, y1, x2, y2]], ...], caption: <description>

Requirements:
- Smooth motion (delta 0.03–0.08 per frame)
- Consistent with phase objectives
- Maintain object sizes
- Only include moving objects: {MOVING_OBJECTS}
\end{verbatim}
\end{tcolorbox}\caption{Prompt used for sub-instruction trajectory planning}
\label{prompt:trajectory-proposal}\end{figure*}

\begin{figure*}\begin{tcolorbox}[
    colback=gray!5,
    colframe=black,
    title={Plan-Alignment Verifier Prompt (Gemini 2.5-Flash)},
    fonttitle=\bfseries,
    arc=2mm
]
\scriptsize
\textbf{System Message:}
\begin{verbatim}
You are an expert at evaluating video motion trajectories.
Your task is to verify if the motion from the first frame to the last frame
aligns with the expected phase goal.

Rate the alignment on a scale of 0.0 to 1.0 where:
- 1.0 = Perfect alignment, the last frame clearly achieves the phase goal
- 0.7-0.9 = Good alignment with minor deviations
- 0.4-0.6 = Partial alignment, some aspects correct
- 0.0-0.3 = Poor alignment, does not achieve the goal

Return ONLY a JSON object with:
{
  "score": <float between 0 and 1>,
  "explanation": "<brief explanation of why this score was given>"
}
\end{verbatim}

\textbf{User Message (paired with first/last-frame images):}
\begin{verbatim}
Phase: {PHASE_NAME}
Phase Description: {PHASE_DESCRIPTION}
Expected End Goal: {END_GOAL}

Please compare the FIRST frame (starting state) with the LAST frame (ending state).
Does the last frame show that the phase goal has been achieved?

Consider:
- Object positions relative to the goal
- Whether objects moved in the expected direction
- Whether the motion is consistent with the phase description
\end{verbatim}
\end{tcolorbox}\caption{Prompt used for plan-alignment verification}
\label{prompt:alignment-verify}\end{figure*}

\begin{figure*}\begin{tcolorbox}[
    colback=gray!5,
    colframe=black,
    title={Motion-Consistency Verifier Prompt (Gemini 2.5-Flash)},
    fonttitle=\bfseries,
    arc=2mm
]
\scriptsize
\textbf{Text Prompt (sent together with the sketch video file):}
\begin{verbatim}
Analyze this video sequence and evaluate whether the motion obeys physical laws.

IMPORTANT NOTE: This video is generated by copy & paste composition - each frame is
created by pasting objects onto a background. Therefore, please focus on evaluating
the movement trajectories and positions of individual objects across frames, not
visual quality, shadows, or composition artifacts.

Consider for each moving object (one of the laws):
Newtonian Consistency: acceleration / deceleration should be physically plausible;
Penetration Violation: objects must not pass through static elements;
Gravitational Coherence: objects should not be floating in the air without 
anything holding it
Deformation Consistency: object size should remain stable unless specified.


Rate the physical plausibility on a scale of 0.0 to 1.0 where:
- 1.0 = Perfectly realistic, obeys this physical laws
- 0.7-0.9 = Mostly realistic with minor issues
- 0.4-0.6 = Some unrealistic aspects but acceptable
- 0.0-0.3 = Highly unrealistic, violates physics (teleportation, impossible speeds, 
etc.)

Return ONLY a JSON object:
{
  "score": <float between 0 and 1>,
  "explanation": "<brief explanation focusing on object movement quality,
                  highlight any physics violations>"
}

Few-shot examples:
{SAMPLED_POSITIVE_AND_NEGATIVE_EXAMPLES}
\end{verbatim}
\end{tcolorbox}\caption{Base prompt template for motion-consistency verification. At inference
time, the few-shot placeholder is populated with positive and negative examples
randomly sampled from the corresponding criterion-specific exemplar pool.}
\label{prompt:physics-verify}\end{figure*}

\end{document}